\documentclass{article} 
\usepackage{iclr2023_conference,times}

\usepackage{amsmath,amsfonts,bm}

\def\figref#1{figure~\ref{#1}}

\def\secref#1{section~\ref{#1}}

\def\eqref#1{equation~\ref{#1}}

\def\1{\bm{1}}

\DeclareMathAlphabet{\mathsfit}{\encodingdefault}{\sfdefault}{m}{sl}
\SetMathAlphabet{\mathsfit}{bold}{\encodingdefault}{\sfdefault}{bx}{n}

\usepackage{url}
\usepackage{graphicx}
\usepackage{booktabs}
\usepackage{animate}
\usepackage{listings}
\usepackage{algorithm}
\usepackage{algpseudocode}
\usepackage[accsupp]{axessibility}  
\newcommand{\nameofmethod}{StoryDiffusion}
\newcommand{\nameofsa}{Consistent Self-Attention}
\newcommand{\nameofmm}{Semantic Motion Predictor}

\renewcommand{\figref}[1]{Fig.~\ref{#1}}
\newcommand{\tabref}[1]{Tab.~\ref{#1}}

\renewcommand{\secref}[1]{Sec.~\ref{#1}}
\newcommand{\myPara}[1]{\vspace{6pt}\noindent\textbf{#1}}
\newcommand{\sArt}{state-of-the-art~}

\usepackage[pagebackref,breaklinks,colorlinks,citecolor=gray]{hyperref}

\title{\nameofmethod{}: Consistent Self-Attention for Long-Range Image and Video Generation}

\author{Yupeng Zhou$^1$~~~Daquan Zhou$^{2*\dagger}$~~~Ming-Ming Cheng$^{1,3}$~~~Jiashi Feng$^2$~~~Qibin Hou$^{1,3*\dagger}$ \\ \\
$^1$ VCIP, School of Computer Science, Nankai University \\
$^2$ ByteDance Inc. \\
$^3$ Nankai International Advanced Research Institute (Shenzhen Futian) \\ 
{$^{*}$Project lead. $^{\dagger}$Corresponding authors.} \\ \\
{Homepage: \url{https://StoryDiffusion.github.io}}
}

\iclrfinalcopy 
\begin{document}

\maketitle

\begin{abstract}
For recent diffusion-based generative models, maintaining consistent content across a series of generated images, especially those containing subjects and complex details, presents a significant challenge.
In this paper, we propose a new way of self-attention calculation, termed \nameofsa{}, that significantly boosts the consistency between the generated images and augments prevalent pre-trained diffusion-based text-to-image models in a zero-shot manner.
To extend our method to long-range video generation, we  further introduce a novel semantic space temporal motion prediction module, named \nameofmm{}. It is trained to estimate the motion conditions between two provided images in the semantic spaces. 
This module converts the generated sequence of images into videos with smooth transitions and consistent subjects that are significantly more stable than the modules based on latent spaces only, especially in the context of long video generation. 
By merging these two novel components, our framework, referred to as \nameofmethod{}, can describe a text-based story with consistent images or videos encompassing a rich variety of contents.
The proposed \nameofmethod{} encompasses pioneering explorations in visual story generation with the presentation of images and videos, which we hope could inspire more research from the aspect of architectural modifications.
%
\end{abstract}

\section{Introduction}
\label{sec:intro}

Recently, diffusion models have been developed rapidly and demonstrated extraordinary potential for content generation, such as images~\citep{rombach2022high, Peebles2022DiT, podell2023sdxl}, 3D objects~\citep{Zeng2022lion,Zhou2021shape3d} and videos~\citep{ho2022imagen, wang2023modelscope}.
With extensive pre-training and advanced architectures, diffusion models show superior performance in generating very high-quality images and videos over previous generative-adversarial network (GAN) based methods \citep{brock2018large}.
However, generating subject-consistent (e.g. characters with consistent identity and attire) images and videos to describe a story is still challenging for existing models.
The commonly used IP-Adapter~\citep{ye2023ip} taking an image as a reference could be used to guide the diffusion process to generate images similar to it.
However, due to the strong guidance, the controllability over the generated content of the text prompts is reduced.
On the other hand, recent state-of-the-art identity preservation methods, such as InstantID~\citep{wang2024instantid} and PhotoMaker~\citep{li2023photomaker}, focus on identity controllability but the consistency of the attires and the scenarios cannot be guaranteed. 
Hence, in this paper, we aim to find a method that can generate images and videos with consistent characters in terms of both identity and attire while maximizing the controllability of the user via text prompts.

\begin{figure}[!t]
\centering
    \begin{minipage}[t]{0.98\textwidth}
    \centering{\it Consistent images generated by \nameofmethod{}}
        \vspace{5pt}
    \end{minipage}
    \begin{minipage}[t]{0.48\textwidth}
    \centering{\it ``Jungle Adventure''}
    \end{minipage}
    \begin{minipage}[t]{0.48\textwidth}
    \centering \it ``The Moon Exploration by Lecun''
    \end{minipage}
    \begin{minipage}[t]{0.985\textwidth}
    \includegraphics[width=\textwidth]{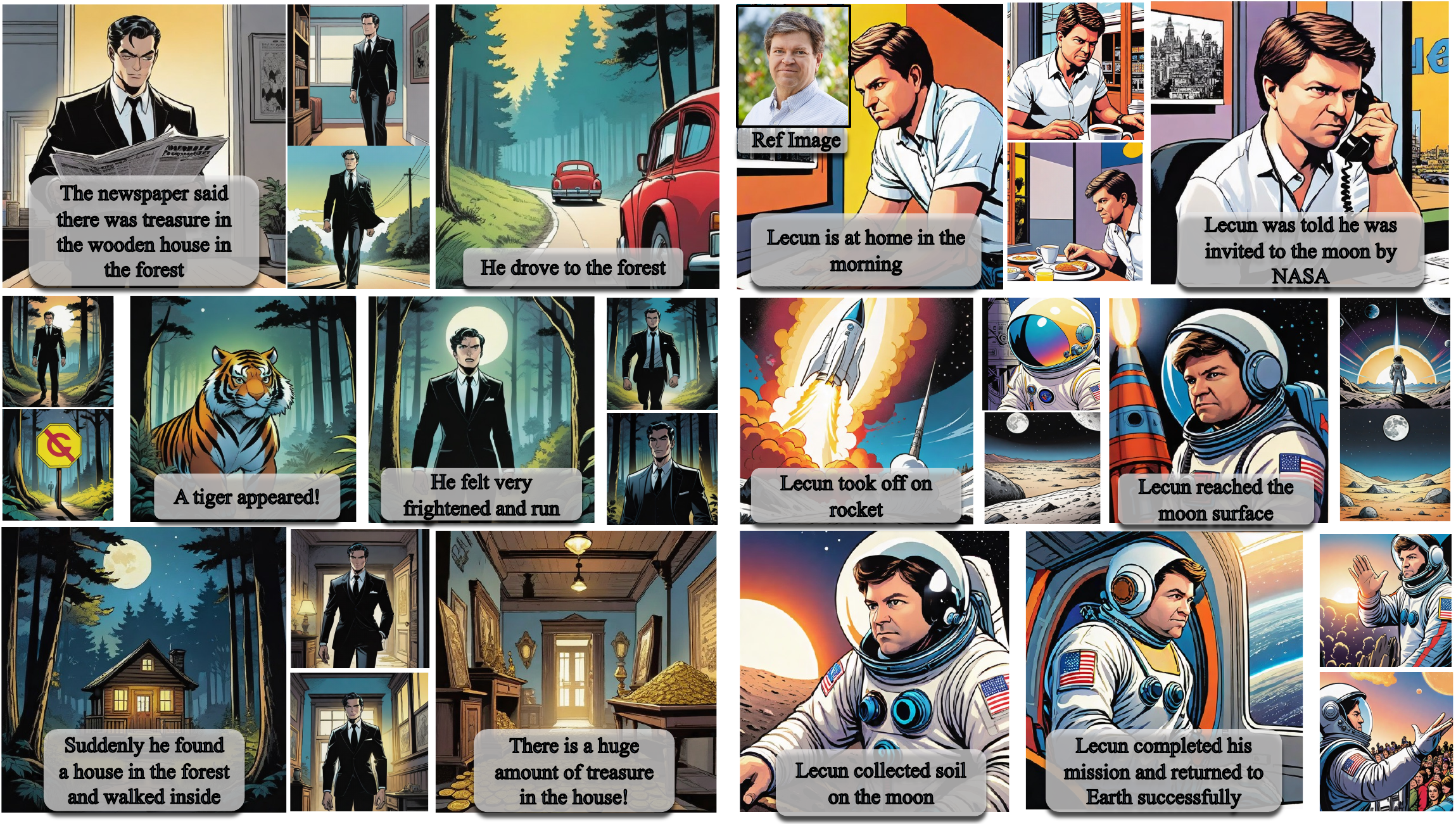}
    \end{minipage}
    \begin{minipage}[t]{0.48\textwidth}
    \centering{(a)}
    \end{minipage}
    \begin{minipage}[t]{0.48\textwidth}
    \centering{(b)}
    \end{minipage}
    \begin{minipage}[t]{0.98\textwidth}
    \centering \it Transition Videos generated by \nameofmethod{}
                \vspace{5pt}
    \end{minipage}
    \begin{minipage}[t]{0.48\textwidth}
    \centering{\it ``Video Clips''}
    \end{minipage}
    \begin{minipage}[t]{0.48\textwidth}
    \centering \it ``Long-Range Video''
    \end{minipage}
    \begin{minipage}[t]{0.24\textwidth}
        \animategraphics[width=\textwidth]{8}{./videos/teaser/ironman1/}{0}{15}
    \end{minipage}
    \begin{minipage}[t]{0.24\textwidth}
        \animategraphics[width=\textwidth]{8}{./videos/teaser/kiss/}{0}{15}
    \end{minipage}
    \begin{minipage}[t]{0.24\textwidth}
        \animategraphics[width=\textwidth]{8}{./videos/teaser/bear/}{1}{77}
    \end{minipage}
    \begin{minipage}[t]{0.24\textwidth}
    \animategraphics[width=\textwidth]{15}{./videos/teaser/man_long/}{60}{212}
    \end{minipage}
    \begin{minipage}[t]{0.98\textwidth}
    \centering{(c)}
    \end{minipage}
    \caption{Images and videos generated by our \nameofmethod{}. (a) Comic generated by \nameofmethod{} telling the story of a man who discovers a treasure while exploring the jungle. (b) Comic generated by \nameofmethod{}  describing the expedition to the moon by Lecun, with a reference image control~\citep{li2023photomaker} same as  \figref{fig:ablation}(b). (c) Videos generated by our \nameofmethod{}. Click the image to play the video. Best viewed with {\it Acrobat Reader}. More generated videos can be found in our home project: \url{https://StoryDiffusion.github.io}.
    } 
    \label{fig:teaser}
\end{figure}

A common approach to preserve the consistency between different images (or frames in the context of video generation) is to use a temporal module~\citep{ho2022video, blattmann2023align}.
However, this requires extensive computational resources and data. Differently, we target to explore a lightweight method with minimum data and computational cost, or even in a zero-shot manner.

As evidenced by previous works~\citep{tian2023diffuse,hong2023improving}, self-attention is one of the most important modules for modeling the overall structure of the generated visual content.
Our main motivation is that if we could use a reference image 
to guide the self-attention calculation, the consistency between the two images is supposed to be improved significantly.
As the self-attention weights are input-dependent, model training or fine-tuning might not be required.  
Following this idea, we propose \nameofsa{}, the core of our \nameofmethod{}, which can be inserted into the diffusion backbone to replace the original self-attention in a zero-shot manner.

Different from the standard self-attention that operates on the tokens representing a single image (as shown in \figref{fig:pipeline}(d)), \nameofsa{} 
incorporates sampled reference tokens from the reference images during the token similarity matrix calculation and token merging.
The sampled tokens share the same set of $Q$-$K$-$V$ weights and thus no extra training is required. 
As shown in \figref{fig:teaser}, the generated images using \nameofsa{} successfully preserve the consistency in both identity and attire, which is vital for storytelling.
Intuitively, \nameofsa{} builds correlations across images in the batch, generating consistent character images in terms of identity and attire, such as clothes.
This enables us to generate subject-consistent images for storytelling.

For any given story text, we begin by dividing it into several prompts, with each prompt corresponding to an individual image.
Then our method could generate highly consistent images that effectively narrate a story.
To support long story generation, we also implement \nameofsa{} together with a sliding window along the temporal dimension. This removes the peak memory consumption's dependency on the input text length, making it possible to generate long stories. 
To stream the generated story frames into videos, we further propose \nameofmm{} that can predict transitions between two images in the semantic spaces. We empirically found that predicting motions in the semantic space generates more stable results than the predictions in the image latent spaces. Combined with the pre-trained motion module~\citep{guo2023animatediff}, \nameofmm{} can generate smooth video frames that are significantly better than recent conditional video generation methods, such as SEINE~\citep{chen2023seine} and SparseCtrl~\citep{guo2023sparsectrl}. 

Our contributions are summarized below:
\begin{itemize}
    \item We propose a training-free and hot-pluggable attention module, termed \nameofsa{}. It can maintain the consistency of characters in a sequence of generated images for storytelling with high text controllability.
    \item We propose a new motion prediction module that can predict transitions between two images in the semantic space, termed \nameofmm{}. It can generate significantly more stable long-range video frames that can be easily upscaled to minutes than recent popular image conditioning methods, such as SEINE~\citep{chen2023seine} and SparseCtrl~\citep{guo2023sparsectrl}.
    \item We demonstrate that our approach could generate long image sequences or videos based on a pre-defined text-based story with the proposed \nameofsa{} and \nameofmm{} with motions specified by text prompts. We term the new framework as \nameofmethod{}.
\end{itemize}

\section{Related Work}

\subsection{Diffusion Models}

Diffusion models have rapidly demonstrated their stunning capabilities in generating realistic images and this also enables them to dominate the field of generative modeling in recent years~\citep{rombach2022high, saharia2022photorealistic, ho2020denoising, ramesh2022hierarchical, Peebles2022DiT}. 
By utilizing a deep denoising network~\citep{ronneberger2015u}, diffusion models establish a connection between the noise distribution and the real image distribution through iterative noise addition and denoising.
Early works~\citep{ho2020denoising, Song2021score, sohl2015deep} establish the theoretical foundation of diffusion models mainly focusing on unconditional image generation. 
Later, various efforts are made to enhance the efficiency and performance of diffusion models.
Typical examples should be efficient sampling methods~\citep{song2020denoising,zhang2022fast, lu2022dpm}, denoising in the latent space~\citep{rombach2022high}, controllability~\citep{feng2022training, zhou2023maskdiffusion}, diffusion backbones~\citep{ramesh2022hierarchical,Peebles2022DiT}.
Concurrently with the exploration of the foundational theory, diffusion models gradually gain popularity and demonstrate strong performance across various domains, such as image generation~\citep{zhou2023maskdiffusion}, video generation~\citep{zhou2023magicvideo}, 3D generation~\citep{Zeng2022lion,Zhou2021shape3d}, image segmentation~\citep{li2023open, amit2021segdiff} and low-level vision tasks~\citep{wang2023exploiting, Lugmayr_2022_CVPR, xie2022smartbrush,saharia2021image}.

\subsection{Controllable Text-to-Image Generation}

As an important sub-field of diffusion model applications, text-to-image generation, represented by Latent Diffusion~\citep{rombach2022high}, DiT~\citep{Peebles2022DiT}, and Stable XL~\citep{podell2023sdxl}, have attracted considerable attention recently.
In addition, to enhance the controllability of text-to-image generation, a multitude of methods emerged as well.
Among them, ControlNet~\citep{zhang2023adding} and T2I-Adapter~\citep{mou2023t2i} introduce control conditions, such as depth maps, pose images, or sketches, to direct the generation of images.
MaskDiffusion~\citep{zhou2023maskdiffusion} and StructureDiffusion~\citep{feng2023trainingfree} focus on enhancing the text controllability.
There are also some works~\citep{mao2023training,ma2023directed} controlling the layout of generated images.

ID-Preservation, which is expected to generate images with a specified ID, is also a hot topic. 
According to whether test-time fine-tuning is required, these works can be divided into two major categories.
The first one only requires fine-tuning a part of the model with a given image, such as Textual Inversion~\citep{gal2022image}, DreamBooth~\citep{ruiz2023dreambooth}, and Custom Diffusion~\citep{kumari2023multi}.
The other one, exemplified by IPAdapter~\citep{ye2023ip} and PhotoMaker~\citep{li2023photomaker}, leverages models that have undergone pre-training on large datasets, allowing the direct use of a given image to control image generation.
Different from both of the two types, we focus on maintaining the subject consistency in multiple images, to narrate a story.
Our \nameofsa{} is training-free and pluggable and can build connections across images within a batch to generate multiple subject-consistent images.

\subsection{Video Generation}

Due to the success of diffusion models in the field of image generation~\citep{rombach2022high, ho2020denoising}, the exploration in the domain of video generation is also becoming popular.
As text is the most intuitive descriptor users can specify, text-based video generation has attracted the most attention~\citep{guo2023animatediff,jiang2023text2performer,singer2022make,wang2023lavie,yang2023probabilistic}.
VDM~\citep{ho2022video} is among the first that extend the 2D U-Net from image diffusion models to a 3D U-Net to achieve video generation.
Due to the significant increase of computational cost for video generation, later works, such as MagicVideo~\citep{zhou2023magicvideo} and Mindscope~\citep{wang2023modelscope}, introduce 1D temporal attention mechanisms, reducing computations by building upon latent diffusion models.
Following Imagen, Imagen Video~\citep{ho2022imagen} employs a cascaded sampling pipeline that generates videos through multiple stages.
Show-1~\citep{zhang2023show} also proposes a multi-stage approach to balance the generation quality and efficiency.

In addition to traditional end-to-end text-to-video (T2V) generation, 
video generation using other conditions is also an important direction.
This type of methods generates videos with other auxiliary controls, such as depth maps~\citep{guo2023sparsectrl,he2023animate}, pose maps~\citep{xu2023magicanimate,hu2023animate,wang2023disco,ma2023follow}, RGB images~\citep{blattmann2023stable,chen2023seine,ni2023conditional}, or other guided motion videos~\citep{zhao2023motiondirector,wu2023lamp}.  
Different from the ambiguity of the text prompt, introducing this conditional information enhances the controllability of video generation.

Our video generation method focuses on transition video generation, which is expected to generate videos with a given start frame and an end frame. 
Typical related works are SEINE~\citep{chen2023seine} and SparseCtrl~\citep{chen2023seine}.
SEINE randomly masks video sequences as the initial input of the video diffusion models in training to enable the predictions of the transition between two frames. 
SparseCtrl introduces a sparse control network to synthesize the corresponding control information for each frame using sparse control data, thereby directing the generation of videos.
However, the aforementioned transition video generation methods rely solely on temporal networks in image latent space for the predictions of intermediate content.
Thus, these methods often perform poorly on complex transitions, such as large-scale movements of characters. 
Our \nameofmethod{} aims to perform predictions in image semantic spaces to achieve better performance and can handle larger movements, which we will show in our experiment section.

\section{Method}
\begin{figure}[!t]
    \centering
    \includegraphics[width=\linewidth]{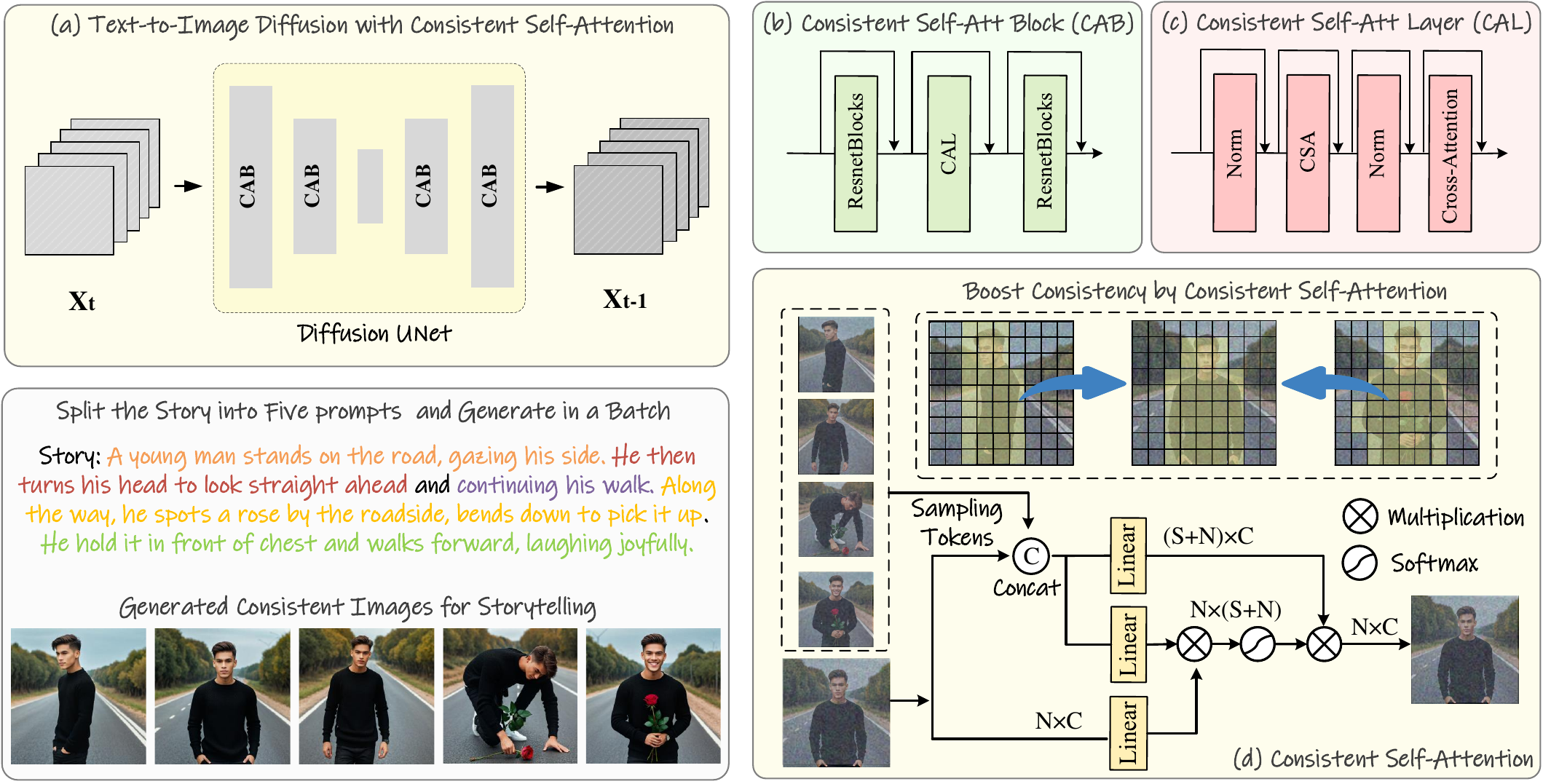}

    \caption{The Pipeline of \nameofmethod{} to generating subject-consistent images. To create subject-consistent images to describe a story, we incorporate our \nameofsa{} into the pre-trained text-to-image diffusion model. We split a story text into several prompts and generate images using these prompts in a batch. \nameofsa{} builds connections among multiple images in a batch for subject consistency.}
    \label{fig:pipeline}
\end{figure}

\begin{figure}[t]
    \centering
    \includegraphics[width=\linewidth]{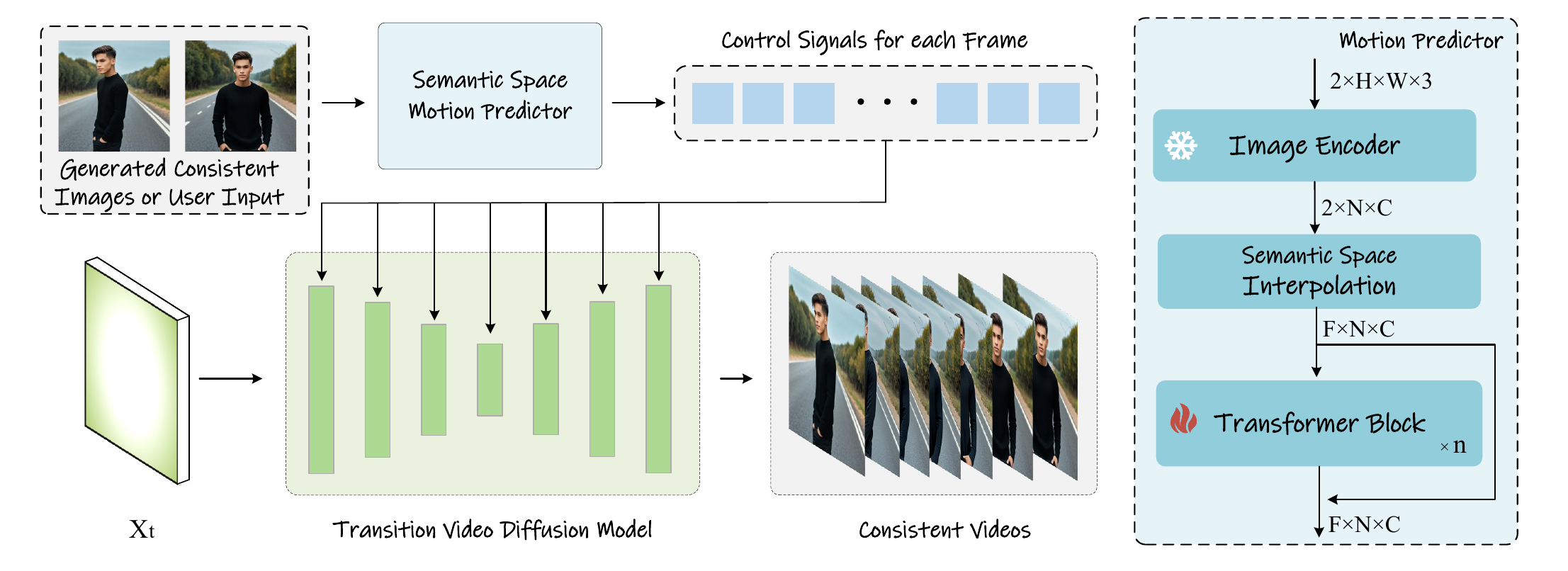}

    \caption{The pipeline of our method for generating transition videos for obtaining subject-consistent images, as described in \secref{sec:31}. To effectively model the character's large motions, we encode the conditional images into the image semantic space for encoding spatial information and predict the transition embeddings. These predicted embeddings are then decoded using the video generation model, with the embeddings serving as control signals in cross-attention to guide the generation of each frame.}
    \label{fig:pipeline2}
\end{figure}

Our method can be divided into two stages, as shown in \figref{fig:pipeline} and \figref{fig:pipeline2}. In the first stage, \nameofmethod{}  utilizes \nameofsa{} to generate subject-consistent images in a training-free manner. These consistent images can be directly applied to storytelling and can also serve as input for the second stage.  In the second stage,   
our \nameofmethod{} create consistent transition videos based on these consistent images.

\subsection{Training-Free Consistent Images Generation}
\label{sec:31}

In this subsection, we introduce how our method generates subject-consistent images in a training-free manner. 
The key to addressing the above issues lies in how to maintain consistency of characters within a batch of images.
This means we need to establish connections between images within a batch during generation.
After revisiting the role of the different attention mechanisms within diffusion models, we get inspired to explore utilizing self-attention to serve the consistency of images within a batch and propose our \nameofsa{}. 
We insert \nameofsa{} into the location of the original self-attention in the existing U-Net architecture of image generation models and reuse the original self-attention weights to remain training-free and pluggable. 

Formally, given a batch of image features $\mathcal{I} \in \mathbb{R}^{B \times N \times C}$, where $B$, $N$, and $C$ are the batch size, number of tokens in each image, and channel number, respectively, we define a function $\operatorname{Attention} ( X_k, X_q, X_v )$ to calculate self-attention. $X_k, X_q$, and $X_v$ stand for the query, key, and value used in attention calculation, respectively. 
The original self-attention is performed within each image feature $I_i$ in $\mathcal{I}$ independently.
The feature $I_i$  is projected to $Q_i$, $K_i$, $V_i$ and sent into the attention function, yielding:
\begin{equation}
    O_i = \operatorname{Attention}\left (Q_i,K_i,V_i\right ).
\end{equation}
To build interactions among the images within a batch to keep subject consistency, our \nameofsa{} samples some tokens $S_i$ from other image features in the batch:
\begin{equation}
    S_i = \operatorname{RandSample}\left (I_1, I_2, .., I_{i-1}, I_{i+1}, ..., I_{B-1}, I_{B}\right ),
\end{equation}
where $\operatorname{RandSample}$ denotes the random sampling function.
After sampling, we pair the sampled tokens $S_i$ and the image feature  $I_i$ to form a new set of tokens $P_i$.
We then perform linear projections on $P_i$ to generate the new key $K_{Pi}$ and value $V_{Pi}$ for \nameofsa{}.
Here, the original query $Q_{i}$ is not changed.
Finally, we compute the self-attention as follows:
\begin{equation}
    O_i = \operatorname{Attention}\left (Q_i, K_{Pi}, V_{Pi}\right ).
\end{equation}

Given the paired tokens, our method performs the self-attention across a batch of images, facilitating interactions among features of different images. This type of interaction promotes the model to the convergence of characters, faces, and attires during the generation process.
Despite the simple and training-free manner, our \nameofsa{} can efficiently generate subject-consistent images, which we will demonstrate in detail in our experiments.
These images serve as illustrations to narrate a complex story as shown in \figref{fig:pipeline}.
To make it clearer, we also show the pseudo code in Algorithm~\ref{alg1}. 

\algrenewcommand\algorithmicindent{0.5em}%
\begin{figure}[t]
\vspace{-20pt}
\begin{algorithm}[H]
\caption{\small Consistent Self-Attention}
\label{alg:train}
\definecolor{codeblue}{rgb}{0.1,0.6,0.1}
\definecolor{codekw}{rgb}{0.85, 0.18, 0.50}
\lstset{
  backgroundcolor=\color{white},
  basicstyle=\fontsize{7.5pt}{7.5pt}\ttfamily\selectfont,
  columns=fullflexible,
  breaklines=true,
  captionpos=b,
  commentstyle=\fontsize{7.5pt}{7.5pt}\color{codeblue},
  keywordstyle=\fontsize{7.5pt}{7.5pt}\color{codekw},
  escapechar={|}, 
}
\begin{lstlisting}[language=python]
def ConsistentSelfAttention(images_features, sampling_rate, tile_size):
  """
  images_tokens: [B, C, N]
  sampling_rate: Float (0-1)
  tile_size: Int
  """
  output = zeros(B, N, C), count = zeros(B, N, C), W = tile_size
  for t in range(0, N - tile_size + 1):
    # Use tile to override out of GPU memory
    tile_features = images_tokens[t:t + W, :, :]
    reshape_featrue = tile_feature.reshape(1, W*N, C).repeat(W, 1, 1)
    sampled_tokens = RandSample(reshape_featrue, rate=sampling_rate, dim=1)
    
    # Concat the tokens from other images with the original tokens
    token_KV = concat([sampled_tokens, tile_features], dim=1)
    token_Q = tile_features
    
    # perform attention calculation:
    X_q, X_k, X_v = Linear_q(token_Q), Linear_k(token_KV), Linear_v(token_KV)
    output[t:t+w, :, :] += Attention(X_q, X_k, X_v)
    count[t:t+w, :, :]  += 1
    
  output = output/count
  return output

\end{lstlisting}
\label{alg1}
\end{algorithm}
\end{figure}

\subsection{\nameofmm{} for Video Generation}
The sequence of the generated character-consistent images can be further refined to videos by inserting frames between each pair of adjacent images. This can be regarded as a video generation task with known start and end frames as conditions. 
However, we empirically observed that recent methods, such as  SparseCtrl~\citep{guo2023sparsectrl} and SEINE~\citep{chen2023seine}, cannot join two condition images stably when the difference between the two images is large. 
We argue that this limitation stems from their sole reliance on temporal modules to predict intermediate frames, which may be not enough to handle the large state gap between the image pair. 
The temporal module operates within pixels on each spatial location independently, therefore, there may be insufficient consideration of spatial information when inferring intermediate frames.
This makes it difficult to model the long-distance and physically meaningful motion.

To address this issue, we propose \nameofmm{}, which encodes the image into the image semantic space to capture the spatial information, achieving more accurate motion prediction from a given start frame and an end frame.
More specifically, in our \nameofmm{}, we first use a function $E$ to establish a mapping from the RGB images to vectors in the image semantic space, encoding the spatial information.
Instead of directly using linear layers as $E$, we utilize a pre-trained CLIP image encoder as $E$ to leverage its zero-shot capabilities for enhancing performance. Using $E$, the given start frame $F_s$ and end frame $F_e$ are compressed to image semantic space vectors  $K_s, K_e$.
\begin{equation}
    K_s,K_e = E\left ( F_s, F_e  \right ).
\end{equation}
Subsequently, in the image semantic space, we train a transformer-based structure predictor to perform predictions of each intermediate frame.
The predictor first performs linear interpolation to expand the two frames $K_s$ and $K_e$ into sequence $K_1, K_2,..., K_L$, where $L$ is the required video length.
Then, the sequence $K_1, K_2,..., K_L$ is sent into a series of transformer blocks $B$ to predict the transition frames:
\begin{equation}
    P_1,P_2,...,P_l = B\left ( K_1, K_2,..., K_L  \right ).
\end{equation}

Next, we need to decode these predicted frames in the image semantic space into the final transition video.
Inspired by the image prompt methods~\citep{ye2023ip}, we position these image semantic embeddings $P_1, P_2, ..., P_L$  as control signals, and the video diffusion model as the decoder to leverage the generative ability of the video diffusion model. 
We also insert additional linear layers to project these embeddings into keys and values, involving into cross-attention of U-Net.

Formally, during the diffusion process, for each video frame feature $V_i$, we concatenate the text embeddings $T$ and the predicted image semantic embeddings $P_i$.
The cross-attention is computed as follows:
\begin{equation}
    V_i = \mathrm{CrossAttention}\left( V_i, \operatorname{concat}(T,P_i),\operatorname{concat}(T,P_i) \right ).
\end{equation}
Similar to previous video generation approaches, we optimize our model by calculating the MSE loss between $L$ frames predicted transition video $O=(O_1, O_2,..., O_L)$ and $L$ frame ground truth $G = (G_1, G_2,..., G_L)$:
\begin{equation}
    Loss = \mathrm{MSE}\left( G,O  \right ).
\end{equation}
By encoding images into an image semantic space for integrating spatial positional relationships,  our \nameofmm{} could better model motion information, enabling the generation of smooth transition videos with large motion.
The results and comparisons that showcase the significant improvements can be observed in \figref{fig:teaser} and \figref{fig:comparsion_video}. 

\section{Experiments}

\subsection{Implementation Details}
For the generation of subject-consistent images, due to the training-free and pluggable property, we implement our method on both Stable Diffusion XL and Stable Diffusion 1.5.
To align with the comparison models, we conduct comparisons on the Stable-XL model using the same pre-trained weights. 
All comparison models utilize 50-step DDIM sampling, and the classifier-free guidance score is consistently set to 5.0.

For the generation of consistent videos, we implement our method based on the Stable Diffusion 1.5 pertained model and incorporate a pretrained temporal module~\citep{guo2023animatediff} to enable video generation. 
All comparison models adopt a 7.5 classifier-free guidance score and 50-step DDIM sampling.
Following the previous methods~\citep{guo2023sparsectrl,chen2023seine}, we use the Webvid10M~\citep{bain2021frozen} dataset to train our transition video model. More details can be found in the supplementary materials.

\subsection{Comparisons of Consistent Images Generation}
\begin{figure}[!t]
    \centering
    \setlength{\abovecaptionskip}{0pt}
    \includegraphics[width=\linewidth]{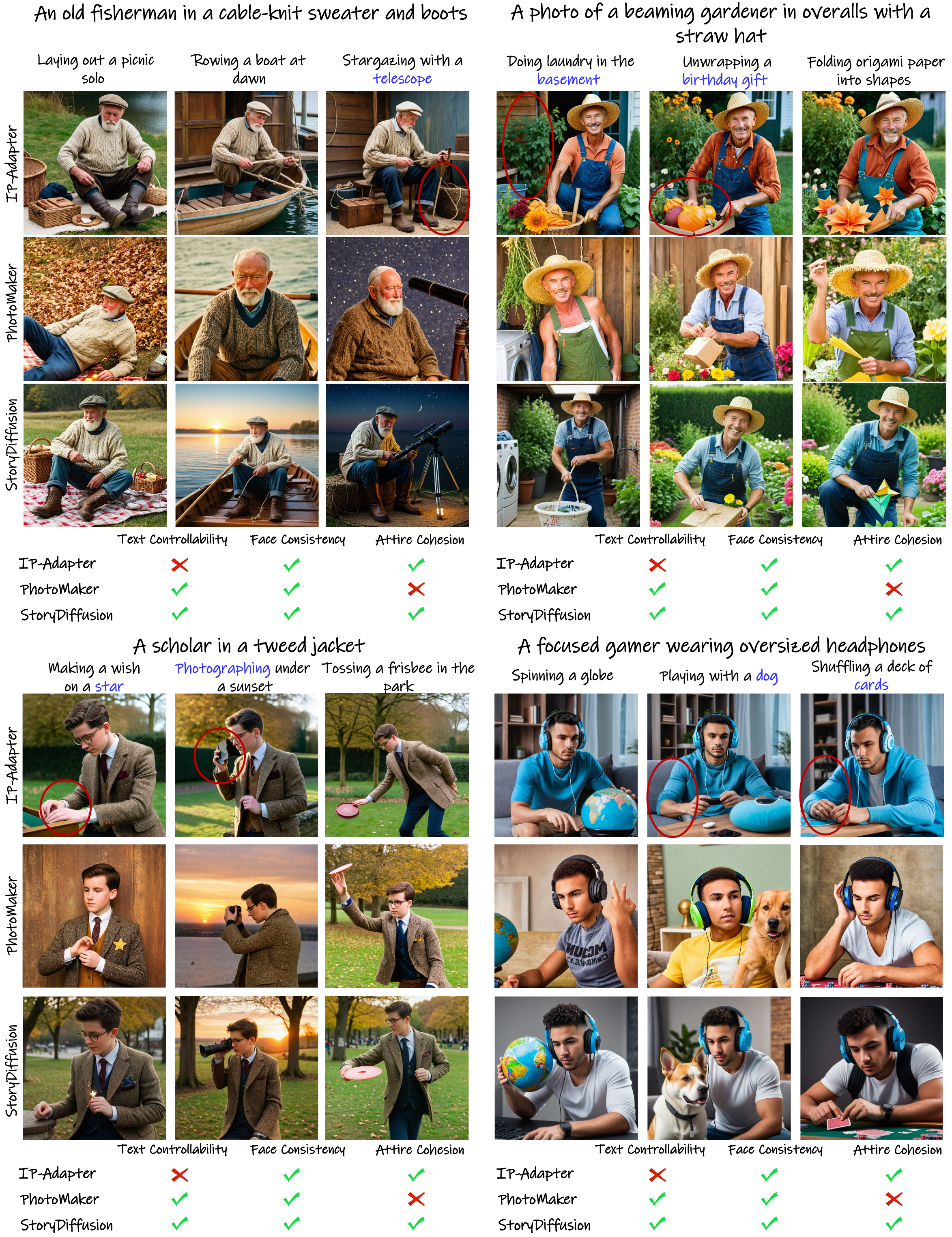}
    \caption{Comparison of consistent image generation with recent methods.}
    \label{fig:comparsion_image}
    \vspace{-15pt}
\end{figure}
We evaluate our method for generating subject-consistent images by comparing it with the two most recent ID preservation methods, IP-Adapter~\citep{ye2023ip} and Photo Maker~\citep{ye2023ip}.
To test the performance, we use GPT-4 to generate twenty character prompts and one hundred activity prompts to describe specific activities. We combine character prompts with activity prompts to obtain 
groups of test prompts.
For each test case, we use the three comparison methods to generate a group of images that depict a person engaging in different activities to test the model's consistency.
Since IP-Adapter and PhotoMaker require an additional image to control the ID of the generated images, we first generate an image of a character to serve as the control image.
We conduct both qualitative and quantitative comparisons to comprehensively evaluate the performance of these methods on consistent image generation.

\begin{table}[ht]
\caption{Quantitative comparisons of consistent image generation. Our \nameofmethod{} achieves better text similarity and subject similarity even without any training.}
\centering
\tabcolsep=4pt
\renewcommand\arraystretch{1.1}
\setlength\arrayrulewidth{1pt}
\scalebox{0.9}{
\begin{tabular}{cccc}
\hline
Metric & IP-Adapter~\citep{ye2023ip} & Photo Maker~\citep{li2023photomaker}  & \nameofmethod{} (ours)  \\ 
\hline
Text-Image Similarity & 0.6129 & 0.6541 & \textbf{0.6586} \\
Character Similarity & 0.8802 & 0.8924 & \textbf{0.8950} \\
\hline
\end{tabular}}
\label{tab:image_quantitative}
\end{table}

\myPara{Qualitative Comparisons.} 
The qualitative result is shown in \figref{fig:comparsion_image}.  
Our \nameofmethod{} can generate highly consistent images, whereas other methods, IP-Adapter and PhotoMaker, may produce images with inconsistent attire or diminished text controllability.
For the first example, the IP-Adapter method generates an image lost ``telescope'' with the text prompt ``Stargazing with a telescope''. PhotoMaker generates images matching the text prompt, but there are significant discrepancies in the attire across the three generated images. The third-row images generated by our \nameofmethod{} exhibit consistent faces and attire with better text controllability. 
For the last example ``A focused gamer wearing oversized headphones'', IP-Adapter lose the ``dog'' in the second image and the ``cards'' in the third image.
The images generated by PhotoMaker could not maintain the attire. Our \nameofmethod{} still generates subject-consistent images, with the same face, and same attire, and conforms to the description in the prompt.

\myPara{Quantitative Comparisons.}  We evaluate the quantitative comparison and show the results in \tabref{tab:image_quantitative}. We evaluate two metrics, the first one is text-image similarity, which calculates the CLIP Score between the text prompts and the corresponding images. The second is character similarity, which measures the CLIP Scores of the character images. Our \nameofmethod{} achieves the best performance on both quantitative metrics, which shows our method's robustness in maintaining character meanwhile conforming to prompt descriptions.

\subsection{Comparisons of Transition Videos Generation}

In transition video generation, we conduct comparisons with the two \sArt methods, SparseCtrl~\citep{guo2023sparsectrl} and SEINE~\citep{chen2023seine}, to evaluate our performance. 
We randomly sample around 1000 videos as the test dataset. 
We employ the three comparison models to predict the intermediate frames of a transition video, given the start and end frames, in order to assess their performance.

\begin{figure}[!htbp]
    \centering
    \includegraphics[width=\linewidth]{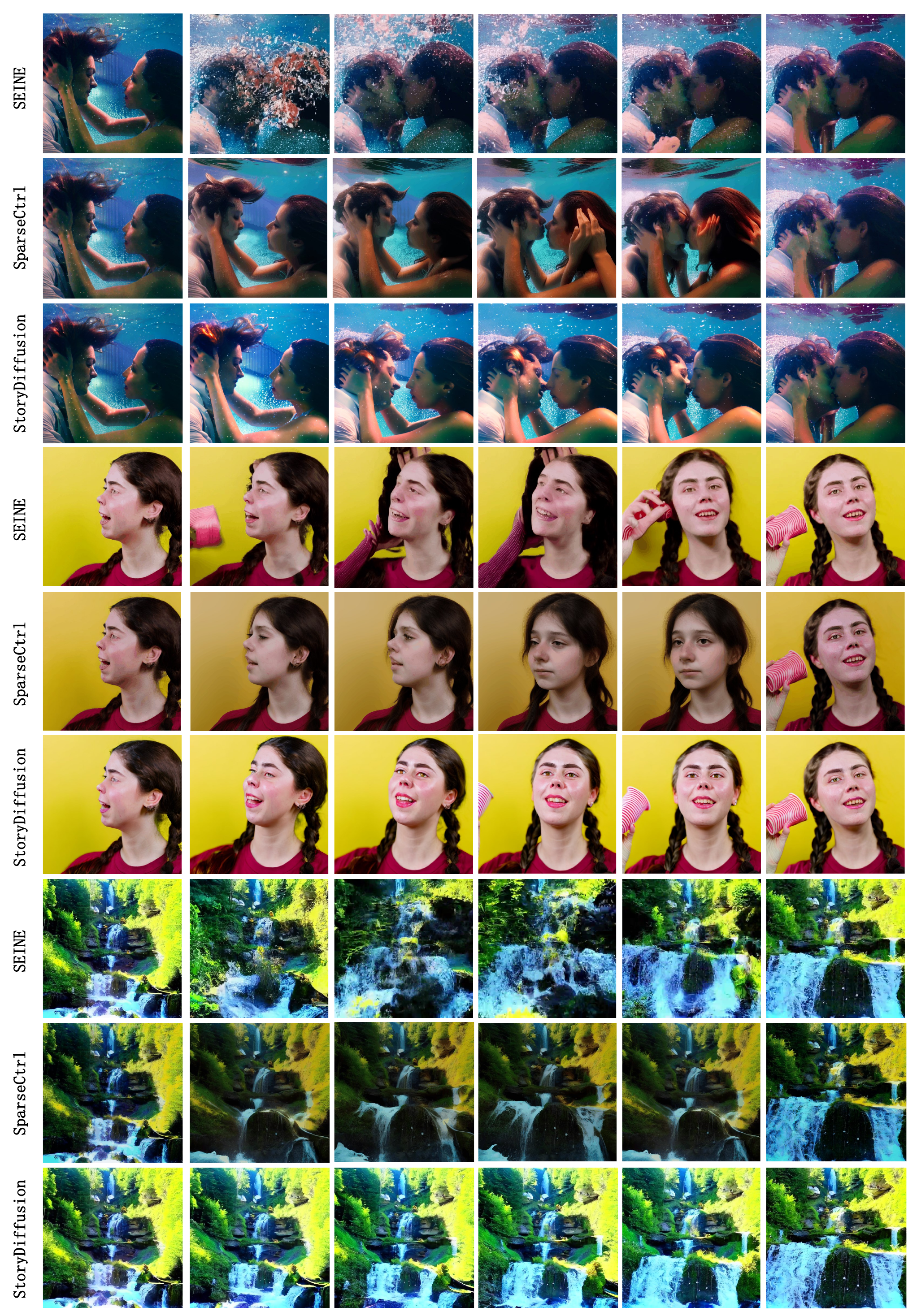}
    \caption{Comparisons of transition video generation with the recent \sArt methods.}
    \label{fig:comparsion_video}
\end{figure}
\myPara{Qualitative Comparisons.} We conduct the qualitative comparison of transition video generation and show the results in \figref{fig:comparsion_video}.
Our \nameofmethod{} significantly outperforms SEINE~\citep{chen2023seine} and SparseCtrl~\citep{guo2023sparsectrl}, generating transition videos that are smooth and physically plausible. 
For the first example, two people kissing underwater, the intermediate frames generated by SEINE are corrupted, and there is a direct jump to the final frame. SparseCtrl generates results with slightly better continuity, but the intermediate frames still contain corrupted images, with numerous hands appearing. However, our \nameofmethod{} succeeds in generating videos with very smooth motion without corrupted intermediate frames. For the second example, the intermediate frames generated by SEINE have corrupted arms. SparseCtrl, on the other hand, fails to maintain consistency in appearance. Our \nameofmethod{} generates consistent videos with excellent continuity. For the last example, the video we generate adheres to physical spatial relationships, unlike SEINE and SparseCtrl, which only change the appearance in the transition.
More visual examples can be found in the supplementary material.

\begin{table}[!t]
\centering
\small
\renewcommand\arraystretch{1.0}
\caption{Quantitative comparisons with \sArt transition video generation models. }
\setlength\arrayrulewidth{1pt}
\tabcolsep=9pt
\begin{tabular}{ccccc}
\hline
Methods  & LPIPS-\textit{first} ($\downarrow$) & LPIPS-\textit{frames} ($\downarrow$) & CLIPSIM-\textit{first} ($\uparrow$)&CLIPSIM-\textit{frames} ($\uparrow$)\\
\cmidrule{1-5}
SEINE 
& 0.4332 & 0.2220 & 0.9259 & 0.9736 \\ 
SparseCtrl 
& 0.4913 & 0.1768 &0.9032 &0.9756 \\
\cmidrule{1-5}
Ours &  \textbf{0.3794} & \textbf{0.1635} & \textbf{0.9606} & \textbf{0.9870}\\
\hline
\end{tabular}   
\vspace{-5pt}
\label{tab:clipsim}
\end{table}

\myPara{Quantitative Comparisons.} Following previous works~\citep{guo2023sparsectrl, zhang2018unreasonable}, we compare our method with SEINE and SparseCtrl with four quantitative metrics, including LPIPS-\textit{first}, LPIPS-\textit{frames}, CLIPSIM-\textit{first}, and CLIPSIM-\textit{frames}, as shown in \tabref{tab:clipsim}. 
LPIPS-\textit{first} and CLIPSIM-\textit{first} measure the similarities between the first frame and other frames, which reflect the overall continuity of the video.
LPIPS-\textit{frames} and CLIPSIM-\textit{frames} measure the average similarities between consecutive frames, which reflect the continuity between frames.
Our model outperforms the other two methods across all four quantitative metrics.
These quantitative experimental results demonstrate the strong performance of our method in generating consistent and seamless transition videos.

\subsection{Ablation Study} 
\begin{figure}[!t]
    \centering
    \includegraphics[width=\linewidth]{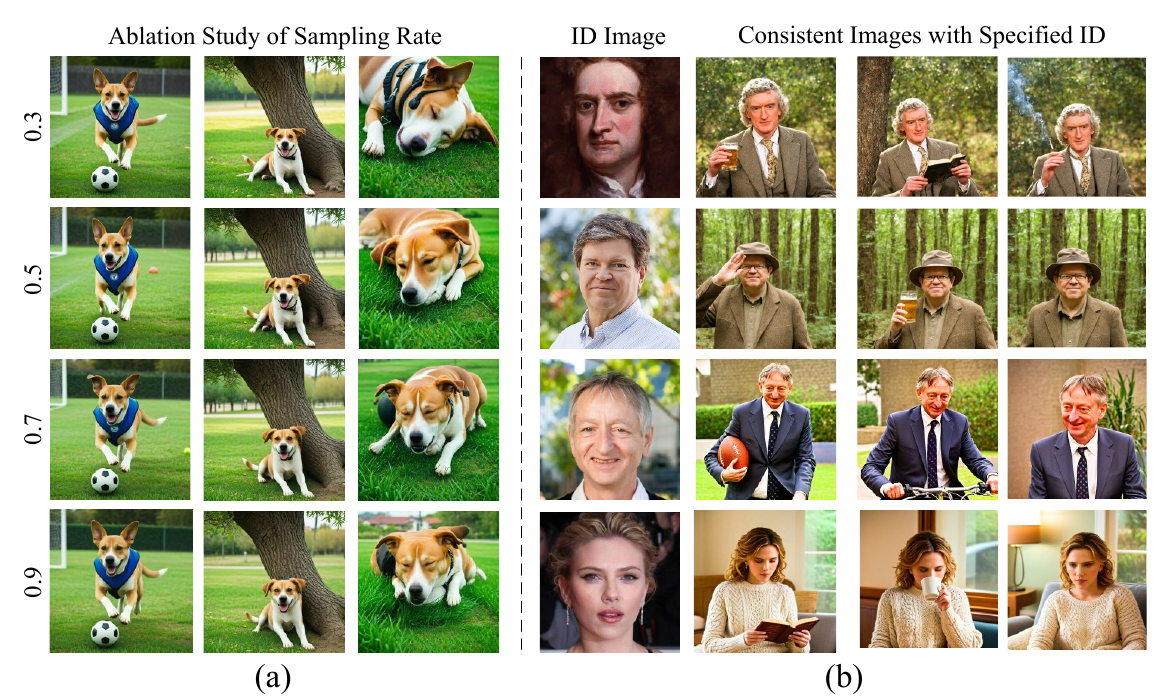}
    \vspace{-10pt}
    \caption{Ablation study. (a)  Evaluations of the impact of different sampling rates in \nameofsa{}. (b) We explore the introduction of external control IDs to govern the generation of characters. Our \nameofmethod{} can generate consistent images that conform to the ID images.  }
    \label{fig:ablation}
\end{figure}

\myPara{User-Specified ID Generation.} We conduct an ablation study to test the performance of generating consistent images with a user-specified ID. Since our \nameofsa{} is pluggable and training-free, we combine our \nameofsa{} with PhotoMaker, giving images to control the characters for consistent image generation.  The results are shown in~\figref{fig:ablation}.
With the control of the ID image, our \nameofmethod{} can still generate consistent images conformed to the given control ID, which strongly indicates the scalability and plug-and-play capability of our method.

\myPara{Sampling Rate of \nameofsa{}.} Our \nameofsa{} sample tokens from other images within a batch and incorporate them into the keys and values during self-attention computation. To determine the optimal sampling rate, we conduct an ablation study on the sampling rate of \nameofsa{}. The results are also shown in~\figref{fig:ablation}. We found a sampling rate of 0.3 could not maintain subject consistency as seen in the third column of images in the left part of ~\figref{fig:ablation}, whereas higher sampling rates successfully preserve it.
In practice, we set the sampling rate to 0.5 by default to make a minimal impact on the diffusion process and maintain consistency.

\begin{table}[!t]
\caption{\textbf{User Study} on subject-consistent image generation and transition video generation.}
\centering
\tabcolsep=14pt
\renewcommand\arraystretch{1.2}
\setlength\arrayrulewidth{1pt}
\small
\begin{tabular}{cccc}
\hline
Consistent Images Generation & IP-Adapter & PhotoMaker  & \nameofmethod{} (ours)  \\ 
\hline
User Preference & 10.4 \% & 16.8 \% & \textbf{72.8 \%}\\
\hline \hline
Transition Video Generation & SEINE & SparseCtrl  & \nameofmethod{} (ours)  \\ 
\hline
User Preference & 11.6 \% & 6.4 \% &  \textbf{82 \%}\\
\hline
\end{tabular}
\label{tab:huamn_eval_overall}
\end{table}

\subsection{User Study}
We conduct a user study with 30 people. Each user is assigned 50 questions to evaluate the effectiveness of our subject-consistent image generation method and transition video generation method. For subject-consistent image generation, we compare with the recent \sArt methods IP-Adapter and PhotoMaker. In transition video generation, we compare with recent \sArt methods SparseCtrl and SEINE.
For fairness, the order of the results is randomized, and users are not informed about which generation model corresponds to each result. The experimental results of our user study are shown in \tabref{tab:huamn_eval_overall}.  Whether for subject-consistent image generation or transition video generation, our model demonstrates an overwhelming advantage. The user study further confirms the superior performance of our \nameofmethod{}.

\section{Conclusions}
In this paper, we propose \nameofmethod{}, a novel method that can generate consistent images in a training-free manner for storytelling and transition these consistent images into videos. Our \nameofsa{} builds connections among multiple images to efficiently generate images with consistent faces and clothing. We further propose the \nameofmm{} to transition these images into videos and better narrate the story. 
We hope that our \nameofmethod{} can inspire future controllable image and video generation endeavors.

\subsubsection*{Acknowledgments}
This research was supported by NSFC (NO. 62225604, No. 62276145), the Fundamental Research Funds for the Central Universities (Nankai University, 070-63223049), CAST through
Young Elite Scientist Sponsorship Program (No. YESS20210377). Computations were supported by the Supercomputing Center of Nankai University (NKSC).

\bibliography{egbib}
\bibliographystyle{iclr2023_conference}

\appendix
\section{Additional Results}

\subsection{Transition Videos generation}
To further showcase the transition video generation capabilities of our \nameofmethod{}, additional videos created by our method are presented in \figref{fig:video}. Our \nameofmethod{} is capable of producing high-quality and smooth transition videos as shown in the additional results.

\begin{figure}[!]
\small
\centering
    \begin{minipage}[t]{0.98\textwidth}
    \centering ``More Transition Videos generated by \nameofmethod{}''
          \vspace{2pt}
    \end{minipage}
    \begin{minipage}[t]{0.20\textwidth}
        \animategraphics[width=\textwidth]{8}{./videos/sup/wine1/}{0}{15}
    \end{minipage}
    \begin{minipage}[t]{0.20\textwidth}
        \animategraphics[width=\textwidth]{8}{./videos/sup/forest1/}{0}{15}
    \end{minipage}
    \begin{minipage}[t]{0.20\textwidth}
        \animategraphics[width=\textwidth]{8}{./videos/sup/man1/}{0}{15}
    \end{minipage}
    \begin{minipage}[t]{0.20\textwidth}
        \animategraphics[width=\textwidth]{8}{./videos/sup/girl5/}{0}{15}
    \end{minipage}
    
    \begin{minipage}[t]{0.20\textwidth}
        \animategraphics[width=\textwidth]{8}{./videos/sup/girl2/}{0}{15}
    \end{minipage}
    \begin{minipage}[t]{0.20\textwidth}
        \animategraphics[width=\textwidth]{8}{./videos/sup/girl1/}{0}{15}
    \end{minipage}
    \begin{minipage}[t]{0.20\textwidth}
        \animategraphics[width=\textwidth]{8}{./videos/teaser/girl3/}{0}{15}
    \end{minipage}
    \begin{minipage}[t]{0.20\textwidth}
        \animategraphics[width=\textwidth]{8}{./videos/sup/boy0/}{0}{15}
    \end{minipage}
    
    \begin{minipage}[t]{0.20\textwidth}
        \animategraphics[width=\textwidth]{8}{./videos/sup/boy3/}{1}{21}
    \end{minipage}
    \begin{minipage}[t]{0.20\textwidth}
        \animategraphics[width=\textwidth]{8}{./videos/sup/man2/}{0}{15}
    \end{minipage}
    \begin{minipage}[t]{0.20\textwidth}
        \animategraphics[width=\textwidth]{8}{./videos/sup/cup0/}{0}{15}
    \end{minipage}
    \begin{minipage}[t]{0.20\textwidth}
        \animategraphics[width=\textwidth]{8}{./videos/teaser/man1/}{1}{15}
    \end{minipage}
    \caption{More video  results generated by our \nameofmethod{}. Best viewed with {\it Acrobat Reader}. Click the image to play the video.}
    \label{fig:video}

\end{figure}

\begin{figure}[!]
    \centering
    \includegraphics[width=0.8\linewidth]{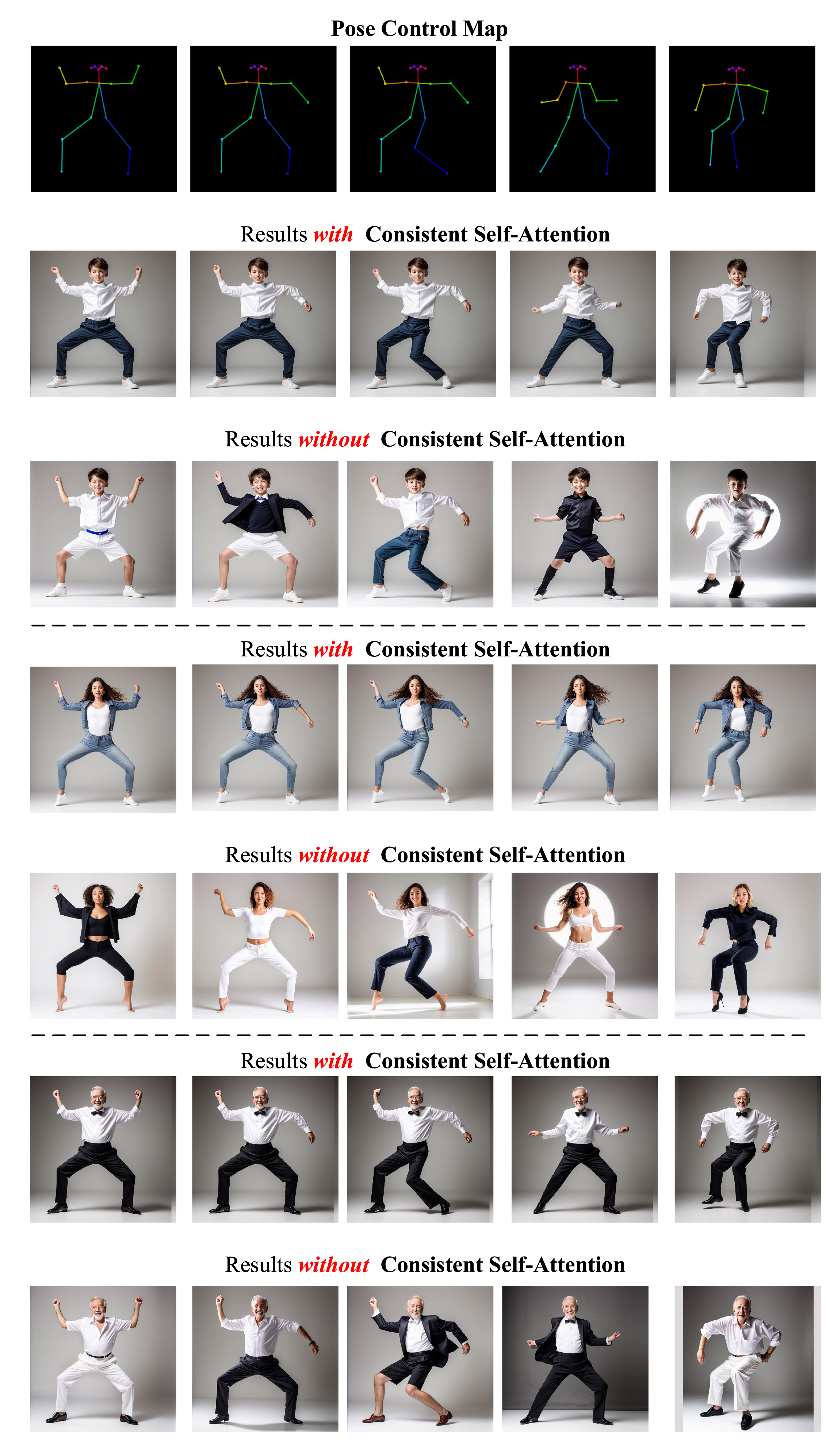}

    \caption{Generation results of our \nameofsa{} combined with ControlNet.}
    \label{fig:controlnet}
\end{figure}

\subsection{Consistent Images Generation with ControlNet}
Given that our \nameofsa{} is training-free and pluggable, we further explore integration with ControlNet~\citep{zhang2023adding} to introduce pose control in the generation of subject-consistent images. The results of combining our \nameofsa{} with ControlNet are displayed in~\figref{fig:controlnet}. Our approach is also capable of generating subject-consistent images under the guidance of ControlNet.

\section{Implementation Details}
For training our transition video model,  we utilize the AnimateDiff V2 motion module~\citep{guo2023animatediff} as our initial weights of the temporal module. We set our learning rate at 1e-4 and conduct training  100k iterations on 8 A100 GPUs.  To encode the conditional images into the image semantic space, we utilize the OpenCLIP ViT-H-14~\citep{radford2021learning,cherti2023reproducible}  pre-trained model.
Our \nameofmm{} incorporates 8 transformer layers, with a hidden dimension of 1024 and 12 attention heads.

\section{Limitations}
The first limitation arises in our subject-consistent image generation.
Similar to current \sArt methods~\citep{ye2023ip}, there may exist inconsistencies in some minor clothing details, such as ties.
In this case, our \nameofsa{} may require more detailed prompts to maintain consistency across images. 
The second limitation is in our transition video generation. Although one can utilize \nameofmethod{} to generate longer videos using a sliding window approach, our method is not designed specifically for long video generation.
Consequently, our method is not yet perfect for generating very long videos due to the absence of global information exchange. 
We will further explore long video generation in our future work.

\end{document}